\documentclass[journal]{IEEEtran}
\usepackage{}
\usepackage{lineno,hyperref}
\usepackage{graphicx}
\usepackage{epstopdf}
\usepackage{mathrsfs}
\usepackage{amsfonts}
\usepackage{bm}
\usepackage{multirow}
\usepackage{threeparttable}
\usepackage[subfigure]{tocloft}
\usepackage{subfigure}

\ifCLASSINFOpdf
\else
\fi
\begin{document}

\title{Robust Face Alignment by Multi-order High-precision Hourglass Network}

\author{Jun~Wan, Zhihui~Lai, Jun~Liu, Jie~Zhou, Can~Gao
	
	
	\thanks{This work is supported by the National Natural Science Foundation of China(Grant No.  62076164, 62002233, 61802267, 61976145 and 61806127), the Natural Science Foundation of Guangdong Province (Grant No. 2019A1515111121, 2018A030310451 and 2018A030310450), the Shenzhen Municipal Science and Technology Innovation Council (Grant No. JCYJ20180305124834854) and the China Postdoctoral Science Fundation (Grant No. 2020M672802). Corresponding author: Zhihui Lai.}
	\thanks{J. Wan is with the College of Computer Science and Software Engineering, Shen zhen University, Shenzhen, 518060, China, and with the Information Systems Technology and Design Pillar, Singapore University of Technology and Design, Singapore, 487372 (e-mail: junwan2014@whu.edu.cn).}
	\thanks{Z. Lai, J. Zhou and C. Gao are with the College of Computer Science and Software Engineering, Shen zhen University, Shenzhen, 518060, China, and the Shenzhen Institute of Artificial Intelligence and Robotics for Society, Shenzhen, 518060, China.(e-mail: lai\_zhi\_hui@163.com, jie\_jpu@163.com, 2005gaocan@163.com)}	
	\thanks{J. Liu is with the Information Systems Technology and Design
		Pillar, Singapore University of Technology and Design, Singapore, 487372 (e-mail: jun\_liu@sutd.edu.sg).}
}

\markboth{Journal of \LaTeX\ Class Files,~Vol.~14, No.~8, August~2015}%
{Shell \MakeLowercase{\textit{et al.}}: Bare Demo of IEEEtran.cls for IEEE Journals}

\maketitle
\begin{abstract}
Heatmap regression (HR) has become one of the mainstream approaches for face alignment and has obtained promising results under constrained environments. However, when a face image suffers from large pose variations, heavy occlusions and complicated illuminations, the performances of HR methods degrade greatly due to the low resolutions of the generated landmark heatmaps and the exclusion of important high-order information that can be used to learn more discriminative features. To address the alignment problem for faces with extremely large poses and heavy occlusions, this paper proposes a heatmap subpixel regression (HSR) method and a multi-order cross geometry-aware (MCG) model, which are seamlessly integrated into a novel multi-order high-precision hourglass network (MHHN). The HSR method is proposed to achieve high-precision landmark detection by a well-designed subpixel detection loss (SDL) and subpixel detection technology (SDT). At the same time, the MCG model is able to use the proposed multi-order cross information to learn more discriminative representations for enhancing facial geometric constraints and context information. To the best of our knowledge, this is the first study to explore heatmap subpixel regression for robust and high-precision face alignment. The experimental results from challenging benchmark datasets demonstrate that our approach outperforms state-of-the-art methods in the literature.
\end{abstract}


\IEEEpeerreviewmaketitle

\section{Introduction}

\IEEEPARstart{F}{ace} alignment, also known as facial landmark detection, refers to locating predefined landmarks (eye corners, nose tip, mouth corners, etc.) of a face. As a typical issue in computer vision, face alignment provides rich geometric information for other face analysis tasks, including facial recognition, face frontalization, human-computer interaction, 3D face reconstruction, etc.
\begin{figure}[t]
	\begin{center}
		\includegraphics[width=0.96\linewidth]{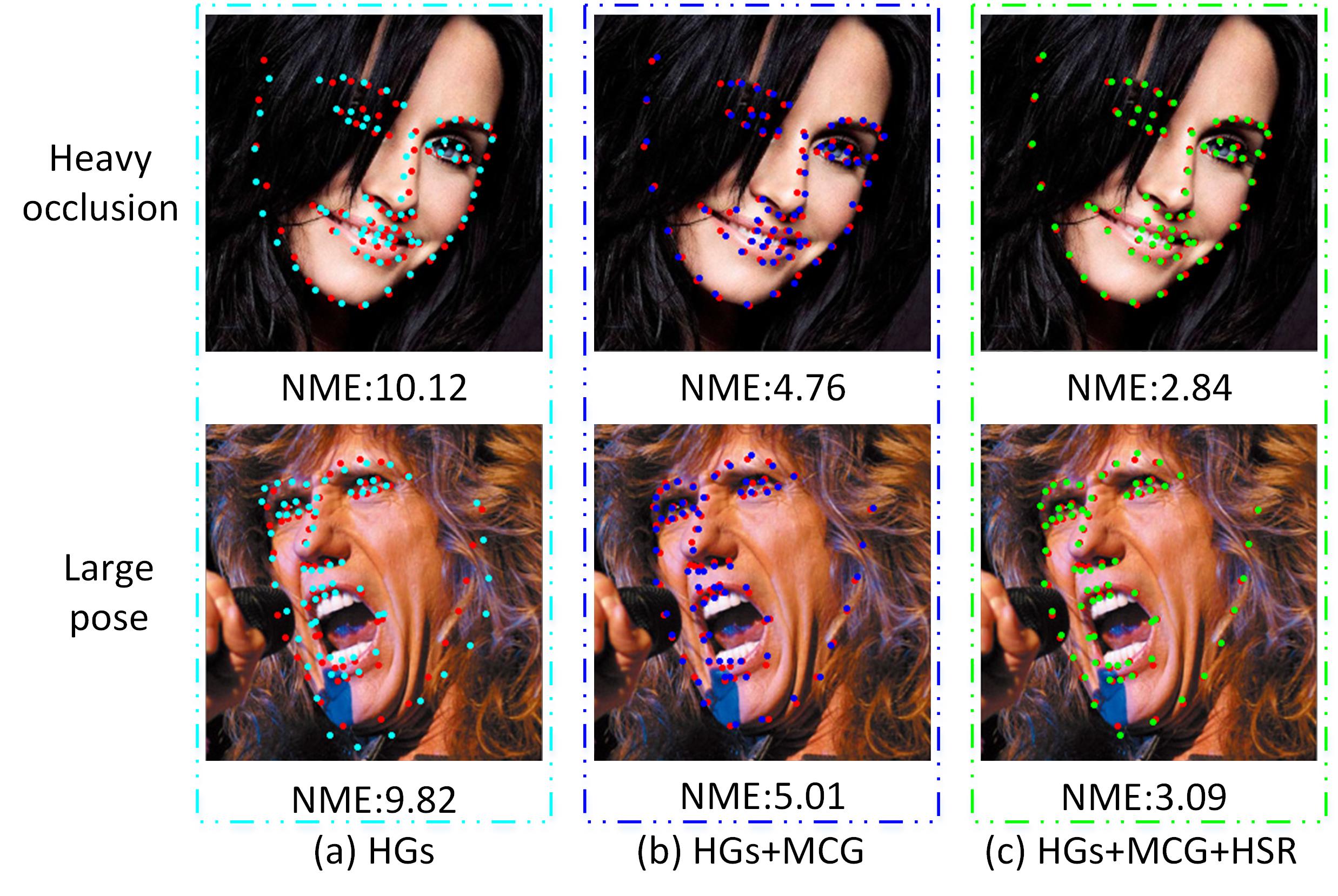}	
	\end{center}
	\caption{(a). The state-of-the-art method HGs\cite{Yang2017StackedHN}; (b). HGs+MCG; (c). HGs+MCG+HSR (MHHN). NME denotes the normalized mean error. The predicted landmarks of HGs, HGs+MCG and HGs+MCG+HSR are denoted by cyan, blue and green colors, respectively, and the ground-truth landmarks are denoted by red color. HGs fails to accurately predict landmarks under heavy occlusions (NME: 10.12) or large poses (NME: 9.82), while the proposed MCG and HSR acheives much better results, i.e., lower NME.}
	\label{figproblem}
\end{figure}
\\\indent In recent decades, scholars have made great progress in face alignment under constrained environments, even by using traditional model-based face alignment algorithms, such as active shape model (ASM) \cite{cootes1995active}, active appearance model (AAM) \cite{cootes2001active}, constrained local model (CLM) \cite{Cristinacce2006FeatureDA} and Gauss-Newton deformable part model (GN-DPM) \cite{Tzimiropoulos2014GaussNewtonDP}. However, the performances of these methods degrade greatly when facing enormous challenges, such as differences in face shapes and facial appearances, i.e., facial expressions, head poses, partial occlusions and illuminations. To enhance the robustness of face alignment in the wild \cite{Jin2017FaceAI, Wang2018FacialFP}, coordinate regression methods have been widely used, in which the mappings from facial appearance features to shape increments are learned by using different kinds of regressors \cite{cao2014face, Merget2018RobustFL, Wu2018LookAB}. Owing to their discriminative features and favorable regression abilities, all these methods are more robust to variations in facial poses and occlusions. However, the coordinate regression method usually regresses to landmark coordinates with a fully connected output layer, which ignores the spatial correlations of features and thus suffers from large poses and partial occlusions. More recently, heatmap regression (HR) methods, such as hourglass networks (HGs \cite{Yang2017StackedHN}), style aggregated network (SAN) \cite{Dong2018StyleAN} and others \cite{Bulat2016HumanPE, Liu2019SemanticAF}, have been proposed as more powerful alternative methods in a wide range of computer vision and pattern recognition tasks, including face alignment. HR methods can effectively drive the model to focus on parts of interest and better encode part constraints and context so that their robustness to large poses and partial occlusions can be enhanced. However, as shown in Fig. \ref{figproblem}, for faces with severe occlusions or extremely large poses, these algorithms fail to accurately predict landmarks because 1) heatmap regression methods are limited by the resolutions of the generated landmark heatmaps and 2) occlusions or large poses may mislead convolutional neural networks (CNNs) in feature representation learning and shape/geometric constraint learning.
\\\indent To address these general problems, in this paper, we propose a novel multi-order high-precision hourglass network (MHHN) (see Fig. \ref{figstructure} for its structure) to achieve heatmap subpixel face alignment by learning more powerful feature representations and feature correlations (i.e., the geometric constraints). Specifically, a subpixel detection loss (SDL) and subpixel detection technology (SDT) are designed to achieve high-precision face alignment via a heatmap subpixel regression (HSR) method. The HSR method can not only help achieve heatmap subpixel-level facial landmark detection but can also make heatmap pixel-level corrections to the detected landmarks. Moreover, we propose a multi-order cross geometry-aware (MCG) model to enhance the feature representation and the geometric constraints by introducing multi-order cross information. The MCG model can also be updated and propagated in stacked hourglass networks, which helps generate more effective landmark heatmaps. Therefore, our method obtains better robustness and accuracy for face alignment under extremely large poses and heavy occlusions.
\\\indent The main contributions of this work are summarized as follows:
\\\indent 1) By incorporating the SDL and SDT, we propose an HSR method that can achieve heatmap subpixel landmark detection. Moreover, the HSR method can be viewed as a generalized framework that can be applied to any heatmap-regression-like tasks such as foreground-background segmentation \cite{Babaee2017ADC}, object segmentation \cite{Yu2015MultiScaleCA}, human pose estimation \cite{Bulat2016HumanPE}, etc.
\\\indent 2) With the well-designed multi-order cross information, a novel MCG model is proposed to enhance the feature representations and geometric constraints for face alignment with extremely large poses and heavy occlusions. 
\\\indent 3) A novel algorithm called MHHN is developed to seamlessly integrate the proposed MCG model and HSR method into a multi-order high-precision hourglass network to handle face alignment under challenging scenarios. To the best of our knowledge, this is the first study to explore heatmap subpixel regression for robust and high-precision face alignment. With the proposed MHHN, our algorithm outperforms state-of-the-art methods on benchmark datasets such as COFW \cite{Burgosartizzu2013Robust}, 300W \cite{Sagonas2016300FI}, AFLW \cite{Zhu2016UnconstrainedFA} and WFLW \cite{Wu2018LookAB}.
\begin{figure*}
	\begin{center}
		\includegraphics[width=0.95\linewidth]{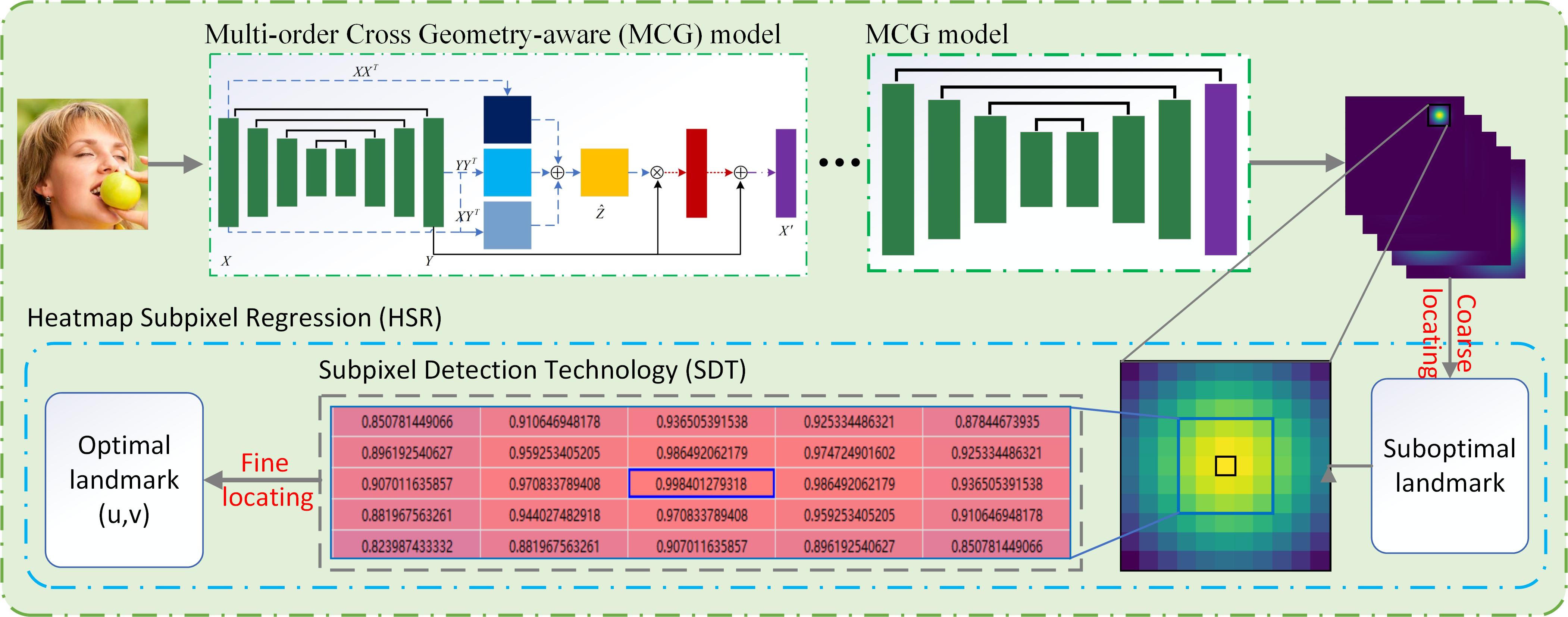}
	\end{center}
	\caption{The proposed multi-order high-precision hourglass network (MHHN). The MCG model is proposed to explore more discriminative representations for enhancing geometric constraints and context information by introducing multi-order cross information, and the HSR method can help to achieve heatmap subpixel landmark detection by combining subpixel detection loss and subpixel detection technology. Then, by integrating the MCG model and HSR method via a seamless formulation, our MHHN is able to achieve more robust FLD.}
	\label{figstructure}
\end{figure*}
\\\indent The rest of the paper is organized as follows. In Section II, we review related works on face alignment. In Section III, we show the proposed method, including the HSR method and MCG model. In Section IV, we conduct a series of experiments to evaluate our proposed method. Finally, we conclude the paper in Section V.
\section{Related Work}
Research on face alignment can be traced back to the 1990s, and since then, rapid development has transpired. In general, existing methods can be categorized into three groups: model-based methods, coordinate regression methods and heatmap regression methods.
\\\indent\textbf{Model-based methods.} The performances of model-based methods  \cite{cootes1995active,cootes2001active, Cristinacce2006FeatureDA, Tzimiropoulos2014GaussNewtonDP} depend on the design of the loss function. AAM \cite{cootes2001active} addresses face alignment by simultaneously matching shape and texture information, which leads to a rapid and accurate algorithm. To reduce the influences of the variations in facial poses and partial occlusions, CLM \cite{Cristinacce2006FeatureDA} first constructs a shape model and a patch model, then searches and matches the predicted landmark around each landmark in the initial shape. GN-DPM \cite{Tzimiropoulos2014GaussNewtonDP} improves performance and reduces the computational cost of the algorithm by jointly optimizing a global shape model and a part-based appearance model with an efficient and robust Gauss-Newton optimization. However, these methods are still sensitive to large poses and partial occlusions.
\\\indent\textbf{Coordinate Regression methods.} This category of methods directly learns the mapping from facial appearance features to the landmark coordinate vectors with different kinds of regressors \cite{cao2014face, ren2014face, zhang2014facial, Wu2018LookAB, Zhu2019RobustFL}. In explicit shape regression (ESR) \cite{cao2014face}, the fern is used to learn a regressor in a cascaded way by minimizing the alignment errors over the training data, and a correlation-based feature selection method is proposed to ensure its accuracy and efficiency. In local binary features (LBF) \cite{ren2014face}, the random forest is used to learn the local binary features and landmark regressor at the same time, and it can achieve 3000FPS in testing. With these effective linear models, robustness and accuracy of face alignment are enhanced. Other methods \cite{zhang2014facial, Wu2018LookAB, Zhu2019RobustFL} use deep models to predict landmark coordinates. In tasks-constrained deep convolutional network (TCDCN) \cite{zhang2014facial}, CNNs are used to construct a shared representation by jointly learning face alignment with subtly correlated tasks (such as appearance attributes, expressions, demographics and head poses), and a task-wise early stopping scheme is proposed to ensure its convergence. In cascaded regression and de-occlusion (CRD) \cite{wan2020robust}, the generative adversarial networks (GANs) are used to locate the facial occlusions and recover the occluded regions. Then the recovered faces can be utilized to improve the robustness of face alignment under occlusions. In look-at-boundary (LAB) \cite{ Wu2018LookAB}, the stacked hourglass network and the GANs are combined to generate facial boundary heatmaps that can effectively help enhance the shape constraints and improve alignment accuracy. In occlusion-adaptive deep network (ODN) \cite{Zhu2019RobustFL}, the Resnet is utilized to construct the geometry-aware module, the distillation module and the low-rank learning module to overcome the occlusion problem in face alignment. Due to their more powerful non-linearity, the performances of coordinate regression methods can be further enhanced. However, coordinate regression methods usually regress landmark coordinates with a fully connected output layer, which ignores the spatial correlations of features. Hence, these methods do not perform as well as heatmap regression models.
\\\indent\textbf{Heatmap regression methods.} HR methods \cite{Yang2017StackedHN, Dong2018StyleAN, Liu2019SemanticAF} directly predict landmark coordinates from the generated landmark heatmaps. Compared to coordinate regression methods, HR methods can better encode the part constraints and context information and effectively drive the network to focus on the important parts in facial landmark detection, thus achieving state-of-the-art accuracy. Yang et al. \cite{Yang2017StackedHN} use a supervised transformation to normalize faces and then a stacked hourglass network to predict landmark heatmaps. By paying more attention to features with high confidence in an explicit manner, the robustness of this method is enhanced. By transforming the original face image into style-aggregated images, style aggregated network (SAN) \cite{Dong2018StyleAN} is able to address the face alignment problems caused by variations in image styles. Liu et al. \cite{Liu2019SemanticAF} propose finding the semantically consistent annotations by a novel latent variable optimization strategy; then, the ground-truth shape can be updated, and the predicted shape becomes more accurate.

Until now, almost all HR methods \cite{Yang2017StackedHN, Dong2018StyleAN, Zhu2019RobustFL} have been limited to the low resolutions of the generated landmark heatmaps, reducing the accuracy of face alignment. Moreover, HR methods often ignore feature inter-dependencies and important higher-order information \cite{Gao2018GlobalSP, Dai2019SecondOrderAN, Wang2019DeepGG, tai2019towards}; thus, they cannot fully explore the discriminative abilities of the features in neural networks. Therefore, we propose a multi-order high-precision hourglass network by exploiting the heatmap subpixel regression method and multi-order cross information for robust face alignment.
\section{Multi-order High-precision Hourglass Network}
In this section, we first elaborate on the proposed HSR method and then present the MCG model. Specifically, the SDL and SDT are proposed to generate more effective landmark heatmaps and achieve high-precision face alignment. Moreover, multi-order cross information is introduced by the MCG model to explore more discriminative features for enhancing facial geometric constraints and context information. Finally, by fusing the MCG model and the HSR method with a seamless formulation via a multi-order high-precision hourglass network, our MHHN can achieve heatmap subpixel accuracy and is more robust to faces with extremely large poses and heavy occlusions.
\subsection{Heatmap Subpixel Regression} 
\begin{figure}[t]
	\begin{center}
		\includegraphics[width=0.95\linewidth]{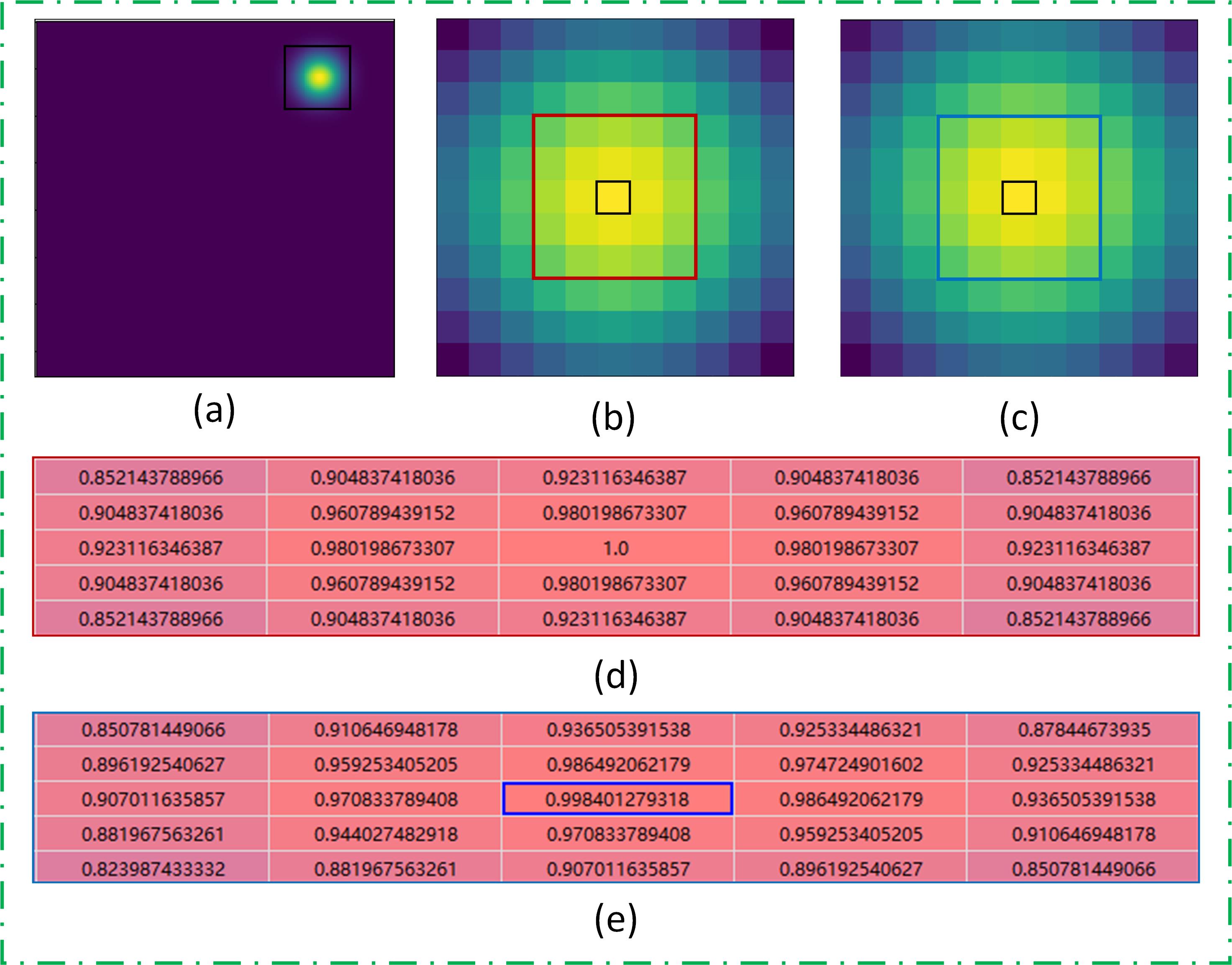}
	\end{center}
	\caption{(a) a landmark heatmap sample. (b) and (c) are the heatmap foreground regions of landmark (24, 24) and (24.2, 23.8), respectively. (d) and (e) are the related digital data of (b) and (c). From (d) and (e), the HR methods can estimate the same landmark coordinates (24, 24). However, the practical landmark in (e) is close to (24, 24) that cannot be detected precisely by the HR methods. Hence, the HR methods are limited to the generated low resolution landmark heatmaps. The practical landmark in (e) can be represented as ($u$, $ v$) which can be accurately estimated by the proposed HSR method.}
	\label{figp2}
\end{figure}

Most HR methods \cite{Yang2017StackedHN, Dong2018StyleAN} utilize the classical (mean squared error) MSE loss to generate landmark heatmaps and then estimate the landmarks by traversing the corresponding landmark heatmaps, which causes the following problems: 1) optimizing the MSE loss to generate landmark heatmaps often makes the generated landmark heatmaps blurry and implausible, reducing the accuracy of face alignment, and 2) traversing the generated landmark heatmaps to predict landmarks limits the accuracy of predicted landmarks to the generated low-resolution heatmaps (see Fig. \ref{figp2}). Transforming the estimated landmarks into the original image size to obtain the final landmark coordinates further increases the error. Hence, in this paper, we propose a heatmap subpixel regression method to handle the above problems. Our HSR method is able to achieve heatmap subpixel accuracy by integrating the SDL and SDT. The proposed SDL mainly contains two parts: the Jensen-Shannon divergence loss and the fine detection loss. The training process of the neural network is as follows. First, by initializing the neural network, the landmark heatmap can be generated. Then, we calculate the Jensen-Shannon divergence loss between the generated and ground-truth landmark heatmaps and predict the heatmap pixel landmark coordinates by traversing the generated heatmaps (corresponding to coarse locating). Next, the heatmap subpixel landmark coordinates can be further estimated by introducing the SDT (corresponding to fine locating), and the fine detection loss can be calculated and regarded as effective feedback for designing the SDL. Finally, by optimizing the SDL, the parameters of the neural networks can be updated. Due to this coarse-to-fine structure, the HSR method can achieve heatmap subpixel landmark detection.
\subsubsection{Jensen-Shannon Divergence Loss}
\indent\indent The ground-truth landmark heatmap for face alignment is generated by a two-dimensional Gaussian distribution, which can be illustrated as:
\begin{equation}
		{H^ * }\left( {x,y} \right) = \exp \left( { - \frac{{{{\left( {x - {u^*}} \right)}^2} + {{\left( {y -{v^*} } \right)}^2}}}{{2{\sigma ^2_*}}}} \right)
\end{equation}where $\left( {{u^*},{v^*}} \right) \in {S^ * }$, $S^ *$ denotes the ground-truth face shape and $ ({u^*},{v^*})$ represents the coordinates of a landmark in $S^ *$. ${H^*}$ is the ground-truth landmark heatmap, while $(x,y)$ is used to denote a pixel's location in heatmap ${H^*}$.
\\\indent In most existing HR methods \cite{Yang2017StackedHN, Dong2018StyleAN}, the MSE loss is utilized to fit neural networks and generate landmark heatmaps. However, the MSE loss treats every location on the heatmap equally, and it is difficult to generate effective landmark heatmaps for high precision facial landmark detection. Here, our proposed HSR method is able to generate more accurate and effective landmark heatmaps because 1) the Jensen-Shannon divergence loss can pay more attention to the foreground area of heatmaps rather than treat the whole heatmap equally; 2) the Jensen-Shannon divergence loss can accurately measure the difference between two distributions and moreover, by optimizing the Jensen-Shannon divergence loss, the generated landmark heatmaps can be further utilized to estimate the heatmap subpixel coordinates. The Jensen-Shannon divergence loss can be stated as:
\begin{equation}
		\begin{array}{l}
			\min JS\left( {\left. {{p_{{H^*}}}} \right\|{p_H}} \right)
			\\ = {\kern 1pt}{\kern 1pt} \;\;\frac{1}{2}KL\left( {{p_H}\left( {x,y} \right)\left\| {\frac{{{p_{{H^*}}}\left( {x,y} \right) + {p_H}\left( {x,y} \right)}}{2}} \right.} \right)\\
		  {\kern 1pt}{\kern 1pt}{\kern 1pt}{\kern 1pt} {\kern 1pt}{\kern 1pt}+ \frac{1}{2}KL\left( {{p_{{H^*}}}\left( {x,y} \right)\left\| {\frac{{{p_{{H^*}}}\left( {x,y} \right) + {p_H}\left( {x,y} \right)}}{2}} \right.} \right)
		\end{array}
\end{equation}where ${p_{{H^*}}}$ and ${p_H}$ denote the distributions of the ground-truth heatmap and the generated heatmap, respectively. ${{p_H}\left( {x,y} \right)}$ represents a pixel value on the heatmap $H$ corresponding to a location $(x,y)$ and works as the confidence of one particular landmark at that pixel. Both ${p_{{H^*}}}$ and ${p_H}$ are based on discrete probability distributions, and we can calculate the Jensen-Shannon divergence loss by a sum operation. By minimizing the Jensen-Shannon divergence loss, we can generate heatmaps with the same distributions and convex points as the ground-truth heatmaps.
\begin{figure*}
	\begin{center}
		\includegraphics[width=0.8\linewidth]{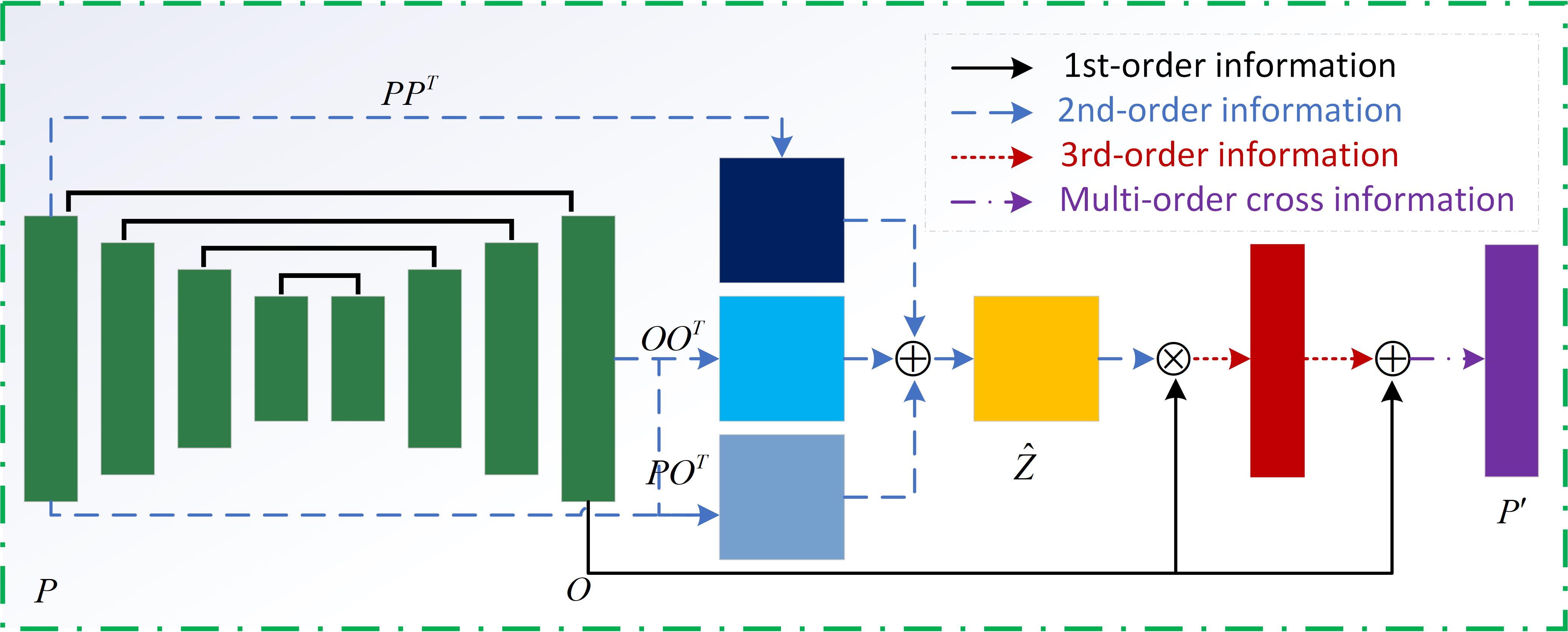}
	\end{center}
	\caption{The network structure of the proposed multi-order cross geometry-aware (MCG) model. With the proposed MCG model, the multi-order cross information containing both cross-layer information (i.e., the autocorrelation of intra-layer features and the cross-correlation of inter-layer features) and cross-order information (the first-order, second-order and third-order information) can be utilized to explore more discriminative representations for enhancing geometric constraints and further generating more effective heatmaps.}
	\label{figmcg}
\end{figure*}
\subsubsection{Subpixel Detection Technology} \indent\indent In existing works \cite{Yang2017StackedHN, Dong2018StyleAN, Zhu2019RobustFL}, by traversing the generated landmark heatmap, the pixel with the maximum value on the heatmap is often regarded as the location of the predicted landmark (the suboptimal landmark). However, these suboptimal landmarks can only achieve heatmap pixel-level accuracy and are not precise enough due to the generated low-resolution landmark heatmaps. Moreover, the transformation of the suboptimal landmarks into the original image size will further increase the errors. Therefore, it is very beneficial to predict the coordinates of the heatmap subpixel landmarks from the generated landmark heatmaps, and we propose the SDT to achieve this. Specifically, SDT first models the continuous distribution of the region centered on the suboptimal landmarks in the heatmaps by using a two-dimensional Gaussian distribution. Then, SDT estimates the parameters of this distribution, and the center of this two-dimensional Gaussian distribution is the predicted heatmap subpixel landmark (the optimal landmark). Because of this continuous distribution, the predicted optimal landmark can achieve heatmap subpixel accuracy, i.e., high-precision facial landmark detection. The modeling of the continuous distribution of the region centered on the suboptimal landmarks in the heatmaps can be formulated as follows:
\begin{equation}
M\left( {x,y} \right) = G\exp \left( { - \frac{{{{\left( {x - u} \right)}^2}}}{{2\sigma _x^2}} - \frac{{{{\left( {y - v} \right)}^2}}}{{2\sigma _y^2}}} \right)
\end{equation}where ${\sigma _x}$ and ${\sigma _y}$ denote the standard deviations in the $x$- and $y$-directions, respectively. $G$ represents the actual maximum confidence in the foreground of the heatmap (not the maximum value obtained by traversing the heatmap). $M$ denotes the confidence matrix of one particular landmark. $(G,{\sigma _x},{\sigma _y},u,v)$ are the parameters of the two-dimensional Gaussian distribution. The solution for $(G,{\sigma _x},{\sigma _y},u,v)$ can be obtained by minimizing the following objective function:
\begin{equation}
		\begin{array}{l}
		\mathop {\min }\limits_{G,{\sigma _x},{\sigma _y},u,v} E\left( {x,y} \right) = \sum\limits_{x,y \in H} {\left\| {M\left( {x,y} \right) - H\left( {x,y} \right)} \right\|_2^2} 
		\end{array}
\end{equation}where $H$ denotes the generated landmark heatmap. Optimization of Eq. (4) can be performed using the least square method or gradient descent method. By minimizing Eq. (4), we can obtain the optimal landmark $\left( {u,v} \right)$.
\subsubsection{Subpixel Detection Loss}
\indent By optimizing the JSDL, landmark heatmaps can be generated, and suboptimal landmarks can be estimated. Then, with the proposed SDT, we can obtain the optimal landmarks and heatmap subpixel face alignment can be achieved. However, optimizing the JSDL or MSE loss leads to less accurate landmark heatmaps, e.g. the maximum value of the generated landmark heatmap is much smaller than 1 or the flat hat problem \cite{Liu2019SemanticAF} arises, which reduces the accuracy of face alignment. Hence, a subpixel detection loss is proposed to address the above problems by combining the Jensen-Shannon divergence loss and fine detection loss. The fine detection loss is defined as follows:
\begin{equation}
		FDL = \sum\limits_k {\left\| {{{\left( {u,v} \right)}_k} - (u^*, v^*)_k} \right\|_2^2}
\end{equation}where $(u^*, v^*)_k$ denotes the coordinates of the  $k$th landmark in the ground-truth face shape ${S^*}$. $(u, v)_k$ represents the  $k$th predicted landmark coordinates by using the SDT. To ensure that the maximum value of the generated landmark heatmap is equal to 1 and solve the flat hat problem \cite{Liu2019SemanticAF}, we set $G=1$ and ${\sigma _x} = {\sigma _y} = {\sigma _*}$. With these settings, the Jensen-Shannon divergence loss and fine detection loss are integrated into the final SDL, which can drive the network to generate more accurate and effective landmark heatmaps and improve the accuracy of face alignment. Moreover, in Table \ref{tabloss}, we also show the corresponding experimental results from the 300W challenging subset without those settings, i.e., we do not specify the values of $G$, ${\sigma _x}$ and ${\sigma _y}$. The SDL is formulated as follows:
\begin{equation}
		\begin{array}{l}
			\mathop {\min }\limits_{W,G,\sigma_x, \sigma_y, u,v} SDL\left( {W,G,\sigma_x, \sigma_y, u,v} \right)  \\ 
			 =\sum\limits_j {\sum\limits_k {JS\left( {p_{H_k^*}^j||p_{{H_k}}^j} \right)} } \\
			  + \lambda \sum\limits_j {\sum\limits_k {\left\| {{{\left( {u,v} \right)}^j_k} - (u^*, v^*)^j_k} \right\|_2^2} }  
		\end{array}
\end{equation}where $j$ denotes the image index in the training set, $k$ represents the landmark index in the face shape and $\lambda $ is the weight coefficient. $W$ represents the parameters of the deep neural networks. With the additional fine detection loss, the SDL can be used to generate more accurate and effective landmark heatmaps. Moreover, the SDT can not only be used to achieve heatmap subpixel-level facial landmark detection but can also make heatmap pixel-level corrections to the detected landmarks. Hence, by integrating the SDL and SDT, heatmap subpixel face alignment can be achieved.
\subsection{Multi-order Cross Geometry-aware Model}
Besides proposing the heatmap subpixel regression method to help achieve heatmap subpixel face alignment, we also explore more effective and discriminative representations for enhancing facial geometric constraints in this paper. HR methods \cite{Yang2017StackedHN, Dong2018StyleAN, Liu2019SemanticAF} achieve state-of-the-art accuracy because they can effectively encode the part constraints and context information. However, these methods suffer from performance degradation for faces with extremely large poses and heavy occlusions, because, in these cases, the extracted features are not robust enough and the facial geometric constraints (e.g. part constraints and global constraints) among landmarks are missing. More recently, second-order information \cite{Gao2018GlobalSP, Dai2019SecondOrderAN, Wang2019DeepGG}, cross-layer features \cite{ long2015fully, kong2016hypernet, cai2016unified, luo2019cross} and feature pyramids \cite{luo2019cross, lin2017feature} have been shown to be useful for obtaining more discriminative and effective representations and are beneficial to many vision tasks. However, how to fuse the higher-order information and cross-layer information to obtain more discriminative representations for robust face alignment is still an open question. Therefore, in this paper, a multi-order cross geometry-aware model is proposed to enhance facial geometric constraints by introducing well-designed multi-order cross information. Multi-order cross information mainly contains cross-order and cross-layer information. Cross-order information can fully explore the useful information contained in features themselves and has a larger receptive field while preserving local details, and cross-layer information can effectively capture landmark geometric constraints at different scales, so multi-order cross information is able to help obtain more discriminative representations for enhancing the geometric constraints, which further helps generate more effective landmark heatmaps and achieve robust face alignment.
\subsubsection{Preliminaries}
The proposed MCG model aims to model the spatial correlations between features (i.e., geometric constraints) by introducing multi-order cross information. As shown in Fig. \ref{figmcg}, the MCG model is a modified hourglass network unit. The input of the hourglass network unit is denoted as $P$, and the output is denoted as $O$. Due to their bottom-up and top-down structures, $P$ and $O$ have the same size, i.e., $P,O \in {\mathbb{R}^{N \times d}}$, where $N = w \times h$, $w$ and $h$ denote the width and height of the feature map, respectively, and $d$ denotes the channel number of the feature map. We use ${P_i}$ to denote the $i$th row of $P$, and ${P_i}$ represents the feature corresponding to the  $i$th location in feature map $P$. The pairwise feature correlations can be represented as their inner product $P_iP_j^T$.
\subsubsection{Construction of the MCG model}
Recently, bilinear CNNs \cite{ lin2017bilinear} and ODN \cite{Zhu2019RobustFL} have used the matrix outer product of the outputs from two CNN streams to model the intrinsic geometric structures of feature maps. Inspired by this work, we use the outer product of $P$ and ${P^T}$ to represent the auto-correlations of intra-layer features that are able to capture geometric correlations between facial landmark regions. $P{O^T}$ is used to represent the cross-correlation of inter-layer features that can capture landmark geometric correlations at different scales (the deep layer and shallow layer). As the outer product operation is similar to a quadratic kernel expansion and is indeed a non-local operation, it can be used to effectively model local pairwise feature correlations for capturing long-range dependencies. Therefore, $P{P^T},O{O^T}$ and $P{O^T}$ can effectively preserve long-range geometric constraints, where $P{O^T}$ denotes the cross-layer information. As shown in Fig. 4, the MCG model computes the sum of $P{O^T},O{O^T}$ and $P{O^T}$:
\begin{equation}
Z = P{P^T} + O{O^T} + P{O^T}
\end{equation}where $Z \in {\mathbb{R}^{N \times N}}$ denotes the pairwise feature correlations of all the pixel locations in the feature maps and $Z$ is the second-order information. The second-order information can fully explore the useful information contained in the features themselves and has a larger receptive field while preserving local details, thus enhancing the geometric constraints of facial landmarks under heavy occlusions. According to \cite{Sun2017HyperlayerBP}, $O{P^T}$ and $P{O^T}$ have similar information, so we discard $O{P^T}$ to reduce the computational cost. 
\\\indent $Z$ is followed by a Softmax function. Building on this, we multiply $\hat Z$ and $O$ to obtain the third-order information, which can be regarded as high-order cross-layer attention feature maps, i.e., we transform the pairwise feature correlations to the high-order cross-layer attention that can be imposed on the original output feature maps $O$. The whole process can be illustrated as follows:
\begin{equation}
\hat Z = softmax\left( Z \right)
\end{equation}
\begin{equation}
P'{\rm{ = }}\gamma \hat ZO{\rm{ + O}}
\end{equation}where $\gamma$ is a learnable scalar that is initialized as 0. $\hat ZO$ and $O$ are the third-order information and first-order information, respectively. Moreover, $\hat Z$ contains the second-order information and cross-layer information. $P'$ is then called the \textbf{multi-order cross information}, i.e., it includes both \textbf{cross-order information} (including first-order, second-order and third-order information) and \textbf{cross-layer information}. Then, by integrating the multi-order cross information into an hourglass network unit, we construct a \textbf{multi-order cross geometry-aware (MCG) model}. Specifically, the multi-order cross information is first introduced into the MCG model, and then, the learnable weight $\gamma$ with respect to the third-order information can be updated gradually during training. Therefore, with the fusion of high-order information and cross-layer information, multi-order cross information can fully explore the useful information contained in feature maps and possesses stronger representation and discrimination capabilities, so it helps the MCG model to better model the facial geometric constraints for robust face alignment with extremely large poses and heavy occlusions.
\\\indent The proposed MCG model has the following three advantages: 1) with the fusion of cross-order information and cross-layer information, the MCG model can obtain more representative and discriminative multi-order cross information; 2) by integrating multi-order cross information into the hourglass network unit, the MCG model can mine useful information inherent in different convolutional layers and different orders of information and 3) multi-order cross information can be updated and propagated in stacked hourglass networks, which can help obtain more accurate and effective landmark geometric constraints for robust face alignment.

\subsection{Multi-order High-precision Hourglass Network}
The proposed MCG model can effectively enhance facial geometric constraints by introducing multi-order cross information, i.e., cross-order information and cross-layer information. Then, the HSR method is proposed to incorporate the well-designed SDL with the SDT for high-precision heatmap regression. Finally, by integrating the MCG model and the HSR method into a novel MHHN via a seamless formulation, we can generate more accurate and effective landmark heatmaps and achieve high-precision face alignment under extremely large poses and heavy occlusions. The overall network structure of the MHHN is shown in Fig. \ref{figstructure}. The objective function of the MHHN can be reformulated as follows:
\begin{equation}
\begin{array}{l}
\mathop {\min }\limits_{W,G,\sigma_x, \sigma_y, u,v} SDL\left( {W,G,\sigma_x, \sigma_y, u,v} \right)\\
= \sum\limits_j {\sum\limits_k {JS\left( {p_{H_k^*}^j||p_{{H_k}}^j} \right)} }  + \lambda \sum\limits_j {\sum\limits_k {\left\| {{{\left( {u,v} \right)}_k} - S_k^*} \right\|_2^2} } \\
= \sum\limits_j {\sum\limits_k {JS\left( {p_{H_k^*}^j||MHHN{{\left( {{I^j},W} \right)}_k}} \right)} } \\
+ \lambda \sum\limits_j {\sum\limits_k {\left\| {{{\left( {u,v} \right)}_k} - (u^*, v^*)^j_k} \right\|_2^2} }
\end{array}
\end{equation}where $MHHN$ denotes the proposed multi-order high-precision hourglass network, and its parameters are denoted by $W$. The input of the MHHN is a face image, and the outputs are landmark heatmaps.
\subsection{Optimization}
The objective function (i.e., Eq. (10)) of the proposed MHHN contains two terms. The first term corresponds to the Jensen-Shannon divergence loss between the predicted and ground-truth landmark heatmaps. The second term corresponds to the fine detection loss, which can be solved by optimizing the following problem:
\begin{equation}
		\mathop {\min }\limits_{G,\sigma_x, \sigma_y, u,v} \sum\limits_{(x,y) \in M} {\left\| {M_k^j\left( {x,y} \right) - MHHN{{\left( {{I^j},W} \right)}_k}} \right\|_2^2}
\end{equation}
\indent Eq. (11) is a classical two-dimensional Gaussian surface fitting problem. Let
\begin{equation}
f\left( {x,y} \right) = G\exp \left( { - \frac{{{{\left( {x - u} \right)}^2}}}{{2\sigma _x^2}} -  - \frac{{{{\left( {y - v} \right)}^2}}}{{2\sigma _y^2}}} \right)
\end{equation}$f \times \ln f$ can be expressed as follows:
\begin{equation}\begin{array}{l}
f \times \ln f = \left( {\ln G - \frac{{{u^2}}}{{2\sigma _x^2}} - \frac{{{v^2}}}{{2\sigma _y^2}}} \right)f + \frac{u}{{\sigma _x^2}}xf\\ \\
{\kern 1pt} {\kern 1pt} {\kern 1pt} {\kern 1pt} {\kern 1pt} {\kern 1pt} {\kern 1pt} {\kern 1pt} {\kern 1pt} {\kern 1pt} {\kern 1pt} {\kern 1pt} {\kern 1pt} {\kern 1pt} {\kern 1pt} {\kern 1pt} {\kern 1pt} {\kern 1pt} {\kern 1pt} {\kern 1pt} {\kern 1pt} {\kern 1pt} {\kern 1pt} {\kern 1pt} {\kern 1pt} {\kern 1pt} {\kern 1pt} {\kern 1pt} {\kern 1pt} {\kern 1pt} {\kern 1pt} {\kern 1pt} {\kern 1pt} {\kern 1pt} {\kern 1pt} {\kern 1pt} {\kern 1pt} {\kern 1pt} {\kern 1pt} {\kern 1pt} {\kern 1pt} {\kern 1pt} {\kern 1pt} {\kern 1pt} {\kern 1pt} {\kern 1pt} {\kern 1pt} {\kern 1pt} {\kern 1pt}  + \frac{v}{{\sigma _y^2}}yf - \frac{u}{{2\sigma _x^2}}{x^2}f - \frac{v}{{2\sigma _y^2}}{y^2}f
\end{array}
\end{equation}
\indent Assuming that there are $L$ data points involved in the fitting, the $L$ data points can be expressed in the form of a matrix, i.e., $A=BC$, where ${a_\ell} = {f_\ell} \times \ln {f_\ell}$, $B = \left[ {{b_\ell}} \right] = \left[{{f_\ell},{f_\ell}{x_\ell},{f_\ell}{y_\ell},{f_\ell}{x_\ell}^2,{f_\ell}{y_\ell}^2} \right]$ and $\ell  = 1,2,...,L$.  $C^T = \left[ {{c_0},{c_1},{c_2},{c_3},{c_4}} \right]$ denotes a vector consisting of Gaussian parameters and can be illustrated as follows:
\begin{equation}
{C^T} = \left[ {\ln G - \frac{{{u^2}}}{{2\sigma _x^2}} - \frac{{{v^2}}}{{2\sigma _y^2}},\frac{u}{{\sigma _x^2}},\frac{v}{{\sigma _y^2}}, - \frac{1}{{2\sigma _x^2}}, - \frac{1}{{2\sigma _y^2}}} \right]
\end{equation}
\indent Then, the least square method is used to fit the errors of the $L$ data points, the fitted model is formulated as follows:
\begin{equation}
\arg \mathop {\min }\limits_C \frac{1}{L}\left\| E \right\|_2^2 = \frac{{{E^T}E}}{L} = \frac{{{{\left( {A - BC} \right)}^T}\left( {A - BC} \right)}}{L}
\end{equation}
\indent With the QR decomposition, $B$ can be reformulated as $B = QR$. $Q$ is an orthogonal matrix and $Q \in \mathbb{R}{^{L \times L}}$. $R$ is an upper triangular matrix and $R \in \mathbb{R}{^{L \times 5}}$. Based on these, Eq. (15) can be reformulated as follows:
\begin{equation}
\begin{array}{l}
\arg \mathop {\min }\limits_C \frac{1}{L}\left\| E \right\|_2^2 = \frac{1}{L}\left\| {{Q^T}E} \right\|_2^2 = \frac{1}{L}\left\| {{Q^T}A - RC} \right\|_2^2\\
\\
\;\;\;\;\;\;\;\;\;\;\;\;\;\;\;\;\;\;\;\;\;\; = \frac{1}{L}\left( {\left\| {S - {R_1}C} \right\|_2^2 + \left\| T \right\|_2^2} \right)
\end{array}
\end{equation}
\begin{equation}
{Q^T}A = \left[ {\begin{array}{*{20}{c}}
	S\\
	T
	\end{array}} \right],{\kern 1pt} {\kern 1pt} {\kern 1pt} {\kern 1pt} R = \left[ {\begin{array}{*{20}{c}}
	{{R_1}}\\
	0
	\end{array}} \right]
\end{equation}where $S$ denotes a 5-dimensional column vector and $T$ represents a  $(L-5)$-dimensional column vector. $R_1$ is an upper triangular matrix and $R_1 \in \mathbb{R}{^{5 \times 5}}$. Hence, Eq. (16) obtains its minimum value when $S=R_1C$, i.e., $C=R_1^{-1}S$. $(u,v)$ can be computed as follows:
\begin{equation}
		u =  - \frac{{{c_1}}}{{2{c_3}}},{\kern 1pt} {\kern 1pt} {\kern 1pt} {\kern 1pt} {\kern 1pt} {\kern 1pt} v =  - \frac{{{c_2}}}{{2{c_4}}}
\end{equation}
\indent For Eq. (12), to ensure that the maximum value of the generated landmark heatmap is equal to 1 and solve the flat hat problem \cite{Liu2019SemanticAF}, we expect to add constraints $G=1$ and ${\sigma _x} = {\sigma _y} = {\sigma _*}$. However, the constraint ${\sigma _x} = {\sigma _y} = {\sigma _*}$ does not always hold, so we firstly only add constraints $G=1$ and $\sigma_x = \sigma_y$, and then calculate $(u,v)$ by further minimizing Eq. (19.1), (19.2) and (19.3), which are shown as follows:
\begin{equation}
	\left\{ {\begin{array}{*{20}{c}}
		{{c_1} - \ln G + \frac{{{u^2} + {v^2}}}{{2\sigma _x^2}}{\kern 1pt} {\kern 1pt} {\kern 1pt} {\kern 1pt} {\kern 1pt} {\kern 1pt} {\kern 1pt} {\kern 1pt} {\kern 1pt} {\kern 1pt} {\kern 1pt} {\kern 1pt} {\kern 1pt} {\kern 1pt} {\kern 1pt} {\kern 1pt} {\kern 1pt} {\kern 1pt} (19.1)}\\
		{{c_2} - \frac{u}{{\sigma _x^2}}{\kern 1pt} {\kern 1pt} {\kern 1pt} {\kern 1pt} {\kern 1pt} {\kern 1pt} {\kern 1pt} {\kern 1pt} {\kern 1pt} {\kern 1pt} {\kern 1pt} {\kern 1pt} {\kern 1pt} {\kern 1pt} {\kern 1pt} {\kern 1pt} {\kern 1pt} {\kern 1pt} {\kern 1pt} {\kern 1pt} {\kern 1pt} {\kern 1pt} {\kern 1pt} {\kern 1pt} {\kern 1pt} {\kern 1pt} {\kern 1pt} {\kern 1pt} {\kern 1pt} {\kern 1pt} {\kern 1pt} {\kern 1pt} {\kern 1pt} {\kern 1pt} {\kern 1pt} {\kern 1pt} {\kern 1pt} {\kern 1pt} {\kern 1pt} {\kern 1pt} {\kern 1pt} {\kern 1pt} {\kern 1pt} {\kern 1pt} {\kern 1pt} {\kern 1pt} {\kern 1pt} {\kern 1pt} {\kern 1pt} {\kern 1pt} {\kern 1pt} {\kern 1pt} {\kern 1pt} {\kern 1pt} {\kern 1pt} {\kern 1pt} {\kern 1pt} {\kern 1pt} {\kern 1pt} {\kern 1pt} {\kern 1pt} {\kern 1pt} (19.2)}\\
		{{c_3} - \frac{v}{{\sigma _x^2}}{\kern 1pt} {\kern 1pt} {\kern 1pt} {\kern 1pt} {\kern 1pt} {\kern 1pt} {\kern 1pt} {\kern 1pt} {\kern 1pt} {\kern 1pt} {\kern 1pt} {\kern 1pt} {\kern 1pt} {\kern 1pt} {\kern 1pt} {\kern 1pt} {\kern 1pt} {\kern 1pt} {\kern 1pt} {\kern 1pt} {\kern 1pt} {\kern 1pt} {\kern 1pt} {\kern 1pt} {\kern 1pt} {\kern 1pt} {\kern 1pt} {\kern 1pt} {\kern 1pt} {\kern 1pt} {\kern 1pt} {\kern 1pt} {\kern 1pt} {\kern 1pt} {\kern 1pt} {\kern 1pt} {\kern 1pt} {\kern 1pt} {\kern 1pt} {\kern 1pt} {\kern 1pt} {\kern 1pt} {\kern 1pt} {\kern 1pt} {\kern 1pt} {\kern 1pt} {\kern 1pt} {\kern 1pt} {\kern 1pt} {\kern 1pt} {\kern 1pt} {\kern 1pt} {\kern 1pt} {\kern 1pt} {\kern 1pt} {\kern 1pt} {\kern 1pt} {\kern 1pt} {\kern 1pt} {\kern 1pt} {\kern 1pt} {\kern 1pt} {\kern 1pt} (19.3)}
		\end{array}} \right.
\end{equation}
\indent The final solution of $(u,v)$ can be expressed as follows:
\begin{equation}
\left\{ {\begin{array}{*{20}{c}}
	{u = \sigma _*^2{c_2},{\kern 1pt} {\kern 1pt} {\kern 1pt} v = \sigma _*^2{c_3}{\kern 1pt} {\kern 1pt} {\kern 1pt} {\kern 1pt} {\kern 1pt} {\kern 1pt} {\kern 1pt} {\kern 1pt} {\kern 1pt} {\kern 1pt} {\kern 1pt} {\kern 1pt} {\kern 1pt} {\kern 1pt} {\kern 1pt} {\kern 1pt} {\kern 1pt} {\kern 1pt} {\kern 1pt} {\kern 1pt} {\kern 1pt} {\kern 1pt} {\kern 1pt} {\kern 1pt} {\kern 1pt} {\kern 1pt} {\kern 1pt} {\kern 1pt} {\kern 1pt} {\kern 1pt} {\kern 1pt} {\kern 1pt} {\kern 1pt} {\kern 1pt} {\kern 1pt} {\kern 1pt} {\kern 1pt} {\kern 1pt} {\kern 1pt} {\kern 1pt} {\kern 1pt} {\kern 1pt} {\kern 1pt} {\kern 1pt} {\kern 1pt} {\kern 1pt} {\kern 1pt} {\kern 1pt}    {\kern 1pt} if{\kern 1pt} {c_1} \ge 0}\\
	{u = \left( {\frac{{ - 2{c_1}}}{{c_2^2 + c_3^2}}} \right){c_2},{\kern 1pt} v = \left( {\frac{{ - 2{c_1}}}{{c_2^2 + c_3^2}}} \right){c_3}{\kern 1pt} {\kern 1pt} {\kern 1pt} {\kern 1pt} {\kern 1pt} {\kern 1pt} {\kern 1pt} {\kern 1pt} {\kern 1pt} {\kern 1pt} {\kern 1pt} {c_1} < 0{\kern 1pt} }
	\end{array}} \right.
\end{equation}
\indent Eq. (20) means that 1) when $c_1 \ge 0$, $G$ equals to 1 and both ${\sigma _x}$ and ${\sigma _y}$ can be set to ${\sigma _*}$ and 2) when $c_1 < 0$, we can only assign $G=1$ and $\sigma_x = \sigma_y={{\frac{{ - 2{c_1}}}{{c_2^2 + c_3^2}}}}$. 

\indent The optimization is a typical network training process under the supervision of the SDL. With the proposed MCG model, we can obtain more representative and discriminative features. By integrating the MCG model and HSR method into the MHHN, the optimal landmark can be easily predicted and heatmap subpixel face alignment is achieved.
\section{Experiments}
\subsection{Datasets}
We evaluate our proposed MHHN method on four challenging datasets: COFW \cite{Burgosartizzu2013Robust}, 300W \cite{Sagonas2016300FI}, AFLW \cite{Zhu2016UnconstrainedFA} and WFLW \cite{Wu2018LookAB}.
\\\indent\textbf{300W} (68 landmarks): With a total of 3148 pictures, the training set is made up of the AFW \cite{Belhumeur2011LocalizingPO}, LFPW \cite{Zhu2012FaceDP} and Helen \cite{le2012interactive} training sets, while the testing set includes 689 images with the IBUG \cite{Sagonas2016300FI}, LFPW and Helen testing sets.
\\\indent\textbf{COFW} (68 landmarks): It contains 1345 training images in which 845 images come from the LFPW \cite{Zhu2012FaceDP} dataset and the other images are heavily occluded. The testing set contains 507 face images with heavy occlusions, large pose variations and expression variations.
\\\indent\textbf{AFLW} (19 landmarks): It contains 25993 face images that are characterized by relatively large differences in poses and expressions. AFLW-full divides the 24386 images into two parts: 20000 for training and 4386 for testing. AFLW-frontal selects 1165 images out of the 4386 testing images to evaluate the alignment algorithm on frontal faces.
\\\indent\textbf{WFLW} (98 landmarks): It contains 10000 faces (7500 for training and 2500 for testing) with 98 landmarks. Apart from landmark annotation, the WFLW also possesses rich attribute annotations (such as occlusions, poses, make-up, illuminations, blurs and expressions) that can be used for a comprehensive analysis of existing algorithms.
\\\indent\textbf{Evaluation metric}: Normalized mean error (NME) is commonly used to evaluate face alignment algorithms. For the 300W, the NME normalized by the inter-pupil distance is used. For the AFLW, we use the NME normalized by the face size given by the AFLW. For the WFLW and COFW, the NME normalized by inter-ocular distance is adopted, which is the same as the evaluation criterion of LAB \cite{Wu2018LookAB}. Moreover, we also use the standard deviation of the NME to evaluate these four datasets.
\\\indent\textbf{Implementation Details}: In our experiments, all the training and testing images are cropped and resized to $256 \times 256$ according to the provided bounding boxes. To perform data augmentation, we randomly sample the angle of rotation and the bounding box scale from a Gaussian distribution. We use four stacked hourglass networks as our backbone to construct the proposed MHHN. During training, we use the staircase function. The initial learning rate is $1 \times {10^{{\rm{ - }}3}}$, which is decayed to $1 \times {10^{{\rm{ - }}5}}$ after 150 epochs. The learning rate is divided by 2, 5, 10 and 10 at epoch 20, 50, 100 and 150, respectively. During the search for the optimal landmark, we predict this landmark from a $9\times9$ region centered on the suboptimal landmark in the generated heatmap. The MHHN is trained with Pytorch on 8 Nvidia Tesla V100 GPUs.
\begin{table}
	\caption{Comparisons with state-of-the-art methods on 300W dataset. The error (NME) normalized by the inter-pupil distance and its standard deviation (in bracket) are given. (- not counted)}
	\begin{center}
		\begin{tabular}{p{3cm}p{1.3cm}p{1.4cm}p{1.3cm}}
			\hline
			Method & Com.  & Chal.  & Full  \\
			\hline
			LBF\cite{ren2014face} &	4.95 (3.97) &	11.98 (7.07)&	6.32 (6.07)\\
			3DDFA(CVPR16)\cite{Zhu2016FaceAA}&	6.15 (5.58)&	10.59 (6.51)&	7.01 (7.97)\\
			RAR(ECCV16)\cite{Xiao2016Robust}&	4.12 (1.86)	&8.35 (4.74)	&4.94 (3.01)\\
			DCFE(ECCV18)\cite{Valle2018ADC}&	3.83 (1.37)&	7.54 (3.47)&	4.55 (2.55)\\
			Wing(CVPR18)\cite{Feng2017WingLF}&	3.27 (0.75)&	7.18 (2.58)&	4.04(1.46)\\
			LAB(CVPR18)\cite{Wu2018LookAB}&	3.42 (1.04)	&6.98 (2.43)&	4.12 (2.01)\\
			SBR(CVPR18)\cite{Dong2018SupervisionbyRegistrationAU}&	3.28 (0.72)&	7.58 (3.58)&	4.10 (1.68)\\
			CRD\cite{wan2020robust}&	-&	6.97 (2.45)&	-\\
			Liu(CVPR19)\cite{Liu2019SemanticAF} &	3.45 (1.02)&	6.38 (1.27)&	4.02 (1.36)\\
			ODN(CVPR19)\cite{Zhu2019RobustFL}	&3.56 (1.07)	&6.67 (1.94)&	4.17 (1.87)\\
			\hline
			HGs&	4.43 (2.24)&	7.56 (3.68)&	5.04 (2.96)\\
			HGs+HSR&	3.69 (1.10)&	6.72 (2.13)&	4.28 (2.22)\\
			HGs+MCG&	3.54 (0.97)&	6.51 (1.46)&	4.12 (1.77)\\
			\textbf{HGs+MCG+HSR(MHHN)}&	\textbf{3.18} (0.61)&	\textbf{6.01} (0.87)&	\textbf{3.74} (0.74)\\
			\hline
		\end{tabular}
	\end{center}
	\label{tab300w}
\end{table}
\subsection{Comparison with state-of-the-art methods}
\begin{figure}[t]
	\begin{center}
		\includegraphics[width=0.90\linewidth]{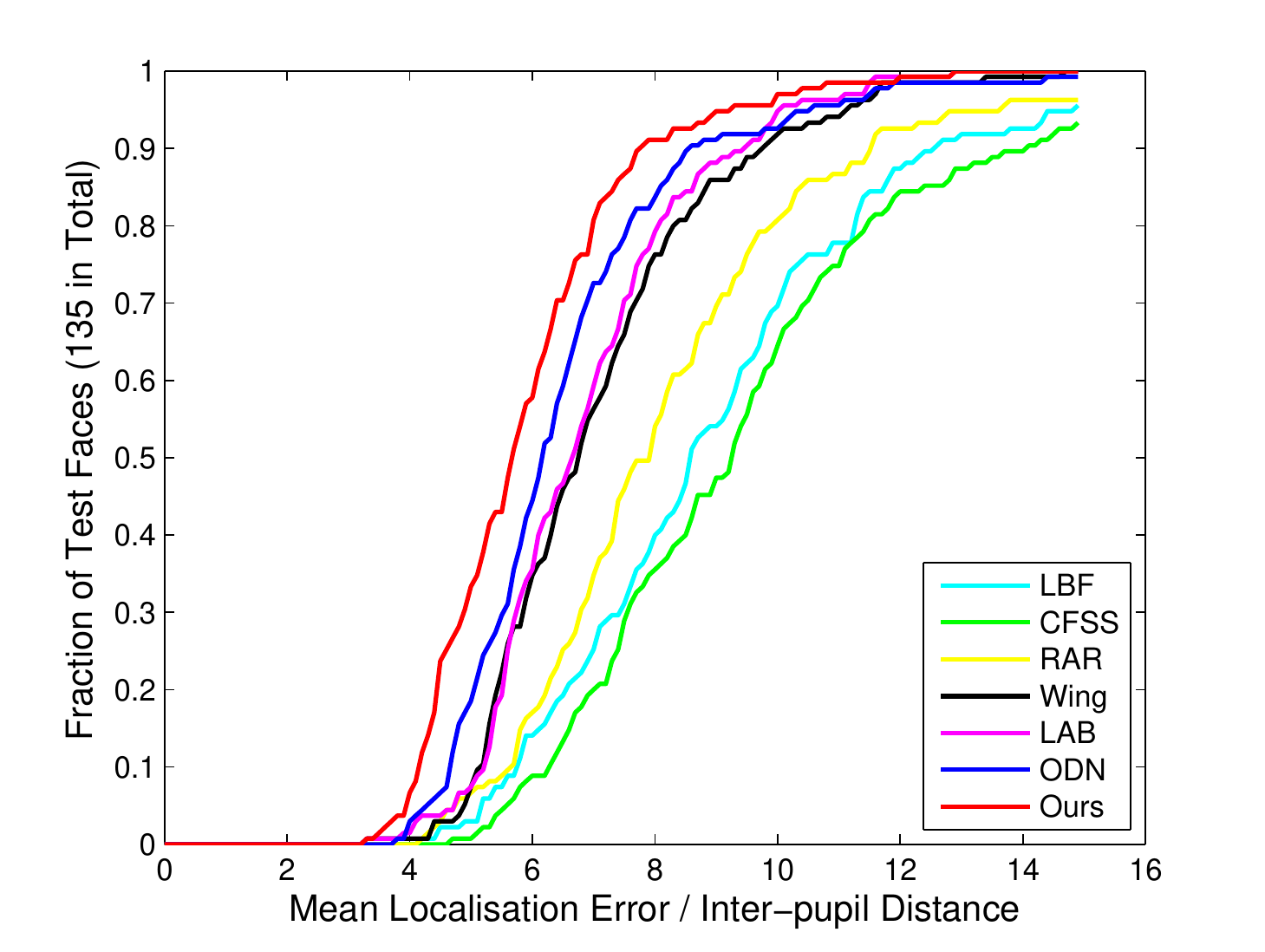}
	\end{center}
	\caption{Comparison of CED curves between our method and state-of-the-art methods including LBF \cite{ren2014face}, CFSS \cite{zhu2015face}, RAR \cite{Xiao2016Robust}, Wing \cite{Feng2017WingLF}, LAB \cite{Wu2018LookAB} and ODN \cite{Zhu2019RobustFL} on 300W Challenging subset (68 landmarks). Our approach is more robust to partial occlusions and large poses than other methods.}
	\label{figchall}
\end{figure}
\begin{figure}[t]
	\begin{center}
		\includegraphics[width=0.96\linewidth]{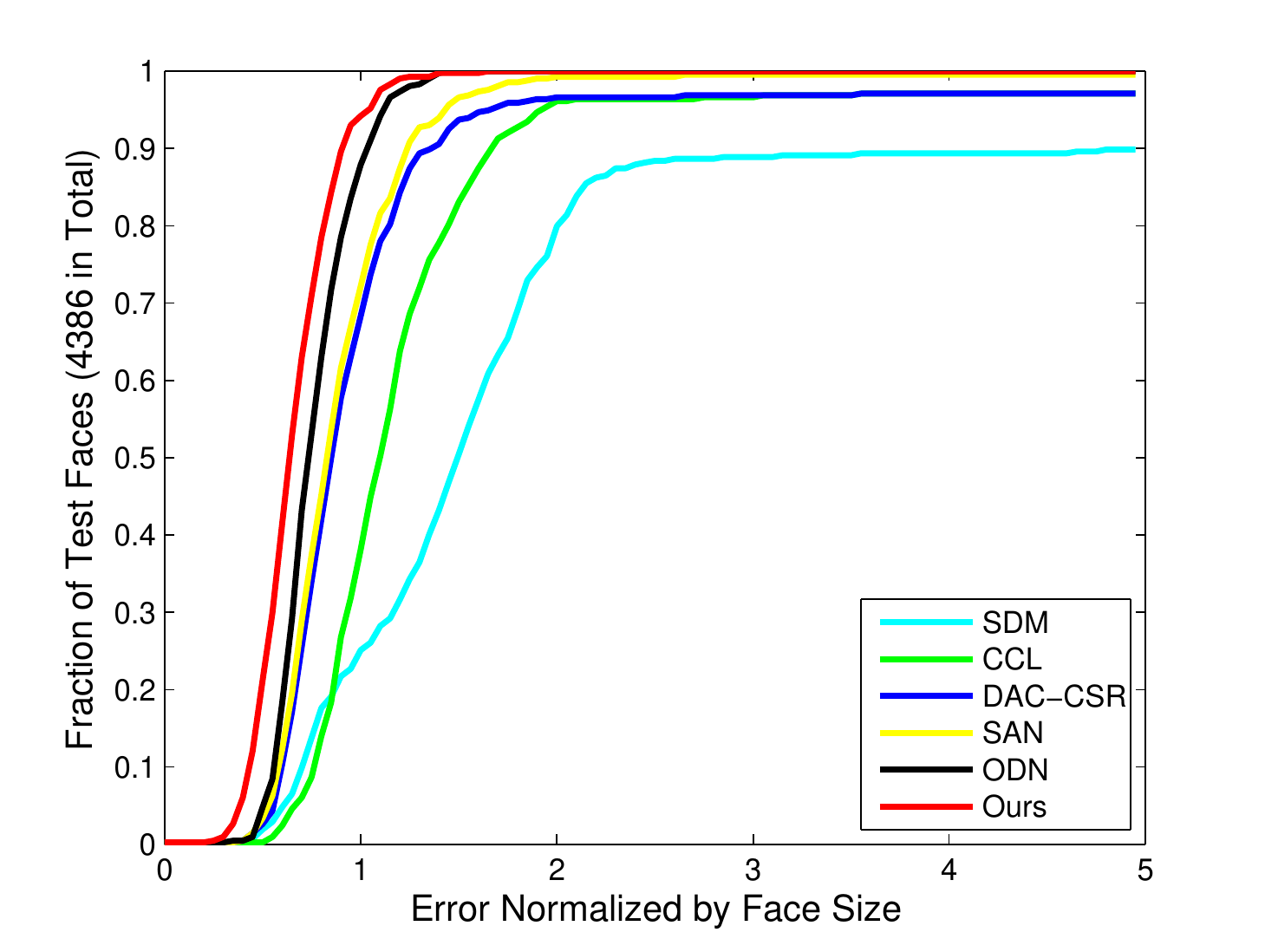}
	\end{center}
	\caption{Comparisons of CED curves of our method and state-of-the-art methods like SDM \cite{xiong2013supervised}, CCL \cite{Zhu2016UnconstrainedFA}, DAC-CSR \cite{Feng2017DynamicAC}, SAN \cite{Dong2018StyleAN} and ODN \cite{Zhu2019RobustFL} on AFLW-full dataset(19 landmarks). Our approach outperforms the other methods.}
	\label{figaflw}
\end{figure}

\textbf{300W.} We compare our approach against state-of-the-art methods on the 300W dataset in Table \ref{tab300w}. The baseline (HGs \cite{Yang2017StackedHN} in Table \ref{tab300w}) uses the original hourglass networks and achieves 4.43\% NME on the 300W Common subset and 7.56\% on the 300W Challenging subset. From Table \ref{tab300w} and Fig. \ref{figchall}, we can see that 1) HGs+HSR achieves 3.69\% NME on the 300W Common subset and 6.72\% NME on the 300W Challenging subset, which are better than HGs alone and indicate that the proposed HSR is helpful for improving the accuracy of face alignment; 2) HGs+MCG achieves 3.54\% NME on the 300W Common subset and 6.51\% NME on the 300W Challenging subset, which also greatly exceed HGs alone and indicate that the proposed MCG model can effectively utilize the multi-order cross information to enhance the geometric constraints and thus generate more robust and effective heatmaps; and 3) HGs+MCG+HSR (MHHN) outperforms state-of-the-art methods, which indicates that the proposed HSR method and MCG model can be seamlessly integrated into a multi-order high-precision hourglass network, thus achieving robust and high-precision face alignment for challenging scenarios.
\begin{table}
	\caption{Comparisons with state-of-the-art methods on COFW dataset. The error (NME) normalized by the inter-ocular distance and its standard deviation (in bracket) are given. (- not counted)}
	\begin{center}
		\begin{tabular}{p{3.8cm}p{1.6cm}p{1.6cm}}
			\hline
			Method & Error  & Failure \\
			\hline
			human&	5.6 (-)&	-\\
			PCPR\cite{Burgosartizzu2013Robust}&	8.50 (7.49)&	20.00\\
			HPM\cite{Ghiasi2014Occlusion}&	7.50 (5.88)&	13.00\\
			CCR\cite{Feng2015Cascaded}&	7.03 (3.53)&	10.9\\
			DRDA\cite{Zhang2016OcclusionFreeFA}&	6.46 (3.21)&	6.00\\
			RAR\cite{Xiao2016Robust}&	6.03 (2.84)&	4.14\\
			DAC-CSR(CVPR17)\cite{Feng2017DynamicAC}&	6.03 (3.07)&	4.73\\
			CAM\cite{Wan2019FaceAB}&	5.95 (2.67)&	3.94\\
			CRD\cite{wan2020robust}&5.72 (2.44) &3.76\\
			LAB(CVPR18)\cite{Wu2018LookAB}&	5.58  (2.17) &	2.76\\
			ODN(CVPR19)\cite{Zhu2019RobustFL}&	5.30  (1.94)&	-\\
			\hline
			HGs&	6.21 (3.29)  &	5.52\\
			HGs+MCG&	5.31 (1.96)  &	3.16\\
			HGs+HSR&	5.62 (2.27)  &	3.75\\	
			\textbf{HGs+MCG+HSR (MHHN)} &	\textbf{4.95} (1.61)&	\textbf{1.78} \\
			\hline
		\end{tabular}
	\end{center}
	\label{tabcofw}
\end{table}
\\\indent\textbf{COFW.} Our method is able to outperform state-of-the-art methods on the COFW dataset, as shown in Table \ref{tabcofw}. The failure rate is defined by the percentage of test images with more than 10\% detection error. As shown in Table \ref{tabcofw}, the accuracy of our proposed HGs+MCG+HSR (MHHN) greatly exceeds that of other methods \cite{Burgosartizzu2013Robust, Feng2015Cascaded, Zhang2016OcclusionFreeFA, Feng2017DynamicAC, Wu2018LookAB, Zhu2019RobustFL}. From the view of another evaluation criterion, the failure rate of our method (1.78\%) is better than those of the other methods. Hence, we can conclude that the proposed method is more robust to heavy occlusions than the other methods.
\\\indent\textbf{AFLW.} Compared with the 300W dataset (68 landmarks), the AFLW has only 19 landmarks, most of which possess challenging shape variations and significant view changes. Table \ref{tabaflw} and Fig. \ref{figaflw} suggest that our method outperforms state-of-the-art methods on the AFLW dataset, indicating that the proposed algorithm has strong robustness over the variations in facial expressions and head poses.
\\\indent\textbf{WFLW.} Compared with other datasets such as the 300W dataset (68 landmarks), AFLW dataset (19 landmarks), WFLW dataset has more landmark annotations, i.e., 98 landmarks. Moreover, images in WFLW dataset are collected from more complicated scenarios. So, experiments on WFLW dataset are more challenging. Table \ref{tabwflw} suggests that our method outperforms state-of-the-art methods on the WFLW dataset, which indicates the proposed HGs+MCG+HSR (MHHN) has strong robustness over the variations in challenging scenarios, i.e., the variations in occlusion, pose, make-up, illumination, blur and expression.
\begin{table}
	\caption{Comparisons with state-of-the-art methods on AFLW dataset. The error (NME) normalized by face size and its standard deviation (in bracket) are given.}
	\begin{center}
		\begin{tabular}{p{3.8cm}p{1.6cm}p{1.6cm}}
			\hline
			Method & full & frontal \\
			\hline
			SDM\cite{xiong2013supervised}&	4.05 (3.91)&	2.94 (3.47)\\
			PCPR\cite{Burgosartizzu2013Robust} & 3.73 (3.68) & 2.87 (3.16)  \\
			ERT\cite{Kazemi2014OneMF}&	4.35 (4.78)&	2.75 (3.21)\\
			LBF\cite{ren2014face} & 4.25 (4.43) & 2.74 (3.12)   \\
			CCL\cite{Zhu2016UnconstrainedFA}&	2.72 (2.41)&	2.17 (2.28)\\
			DAC-CSR(CVPR17)\cite{Feng2017DynamicAC}&	2.27 (2.25)&	1.81 (1.69)\\
			SAN\cite{Dong2018StyleAN}(CVPR18)&	1.91 (1.94)&	1.85 (1.77)\\
			CRD\cite{wan2020robust}&1.80 (1.78) &1.59 (1.29)\\
			ODN\cite{Zhu2019RobustFL}(CVPR19)&	1.63 (1.34)&	1.38 (1.21)\\
			\hline
				HGs&	2.47 (2.33)  &	1.92 (2.01)\\
			HGs+MCG&	1.59 (1.29)  &	1.31 (1.01)\\
			HGs+HSR&	1.76 (1.53)  &	1.48 (1.37)\\	
			\textbf{HGs+MCG+HSR(MHHN)} &	\textbf{1.38} (1.17)&	\textbf{1.19} (0.81) \\
			\hline
		\end{tabular}
	\end{center}
	\label{tabaflw}
\end{table}
\subsection{Self Evaluations}
\begin{figure*}
	\begin{center}
		\includegraphics[width=0.90\linewidth]{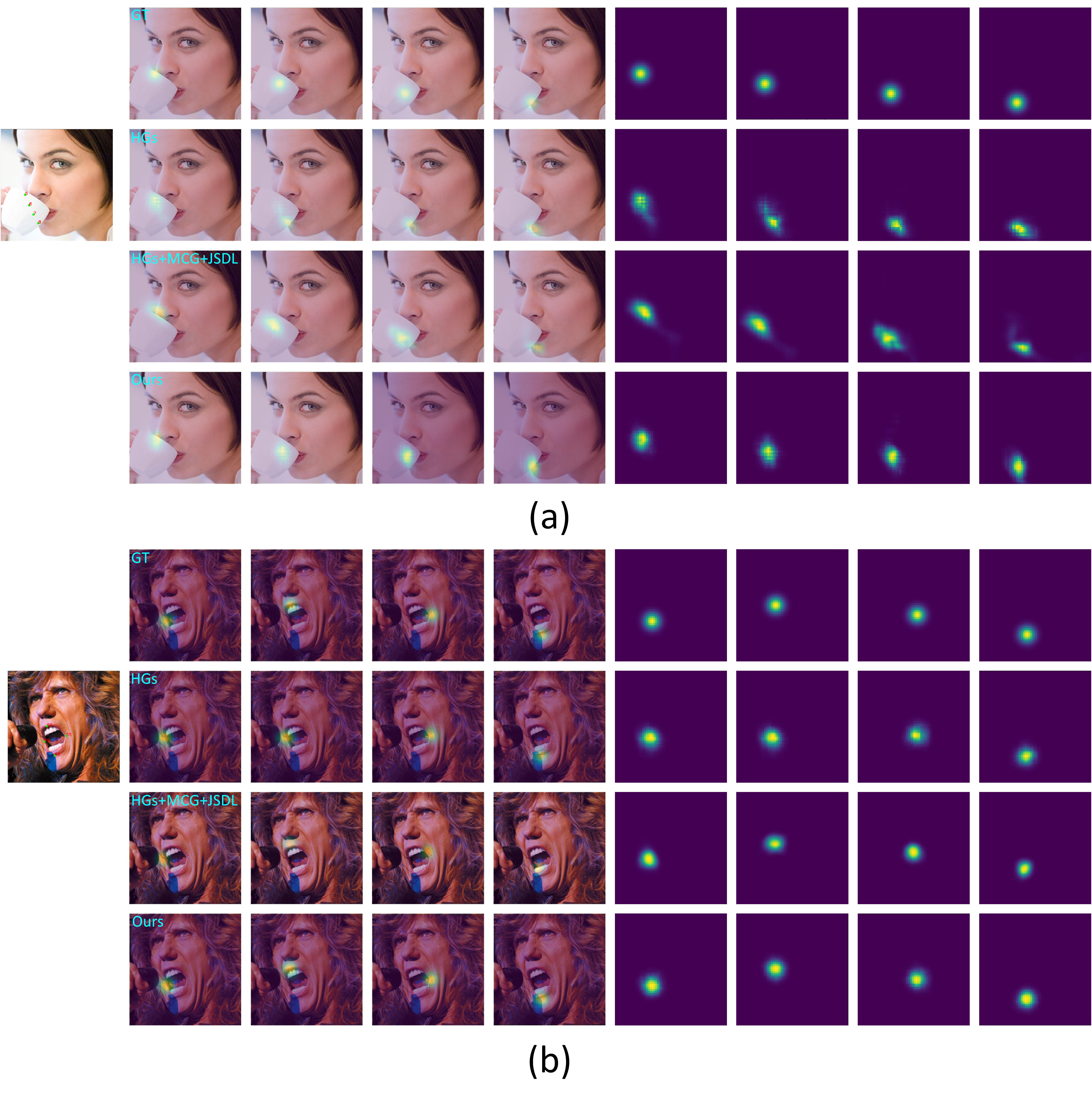}
	\end{center}
	\caption{Comparisons of generated heatmaps of GT (ground-truth), HGs, HGs+MCG+JSDL and our method. With the proposed MHHN, we can find that our heatmaps are more robust to partial occlusions (a) and large poses (b).}
	\label{fig6}
\end{figure*}

\textbf{Loss function.} To generate heatmaps with the same distributions and convex points as the ground-truth heatmaps, we conduct experiments on the 300W challenging subset by separately using the mean squared error (MSE) loss, the Kullback-Leibler divergence loss (KLDL), the Jensen-Shannon divergence loss (JSDL) and SDL. MSE involves applying the mean squared error loss to generated landmark heatmaps and then traversing the landmark heatmaps to obtain the corresponding landmarks. MSE+SDT requires first applying the mean squared error loss to generate landmark heatmaps and then using the SDT to predict the optimal landmarks as a post-processing method. SDL represents the subpixel detection loss with constraints $G=1$ and ${\sigma _x} = {\sigma _y} = {\sigma _*}$, and SDL1 denotes the subpixel detection loss without those constraints. From the experimental results in Table \ref{tabloss}, we can conclude that 1) the SDT can be used to improve the accuracy of heatmap regression face alignment as a post-processing method; 2) compared to the MSE loss and KLDL, the JSDL can help generate more effective heatmaps, which can be combined with the SDT to further improve the performance of the face alignment. 3) by integrating the JSDL and fine detection loss into the well-designed SDL via the proposed HSR method, heatmap subpixel face alignment can be achieved; and 4) by constraining $G=1$ and ${\sigma _x} = {\sigma _y} = {\sigma _*}$, the SDL can generate more accurate landmark heatmaps, and the performance of face alignment can be further improved.
\begin{table*}
	\caption{Comparisons with state-of-the-art methods on WFLW dataset. The error (NME) normalized by the inter-ocular distance and its standard deviation (in bracket) are given.}
	\begin{center}
		\begin{tabular}{p{3cm}p{1.5cm}p{1.6cm}p{1.5cm}p{1.6cm}p{1.6cm}p{1.5cm}p{1.5cm}}
			\hline
			Method & Testset  & Pose Subset  & Expression Subset &Illumination Subset &Make-Up Subset &Occlusion Subset & Blur Subset  \\
			\hline
			ESR\cite{cao2014face} &11.13 (7.91) &25.88 (15.89) &11.47 (8.09) &10.49 (10.28) &11.05 (10.88) &13.75 (9.54) &12.20 (9.33) \\
			SDM\cite{xiong2013supervised} &10.29 (7.58) &24.10 (14.33) &11.45 (8.12) &9.32 (9.17) &9.38 (9.54) &13.03 (9.10) &11.28 (8.78) \\
			CCFS(CVPR15)\cite{zhu2015face} &9.07 (6.87) &21.36 (10.97) &10.09 (7.47) &8.30 (8.22) &8.74 (8.65) &11.76 (8.47) &9.96 (7.89) \\
			DVLN(CVPR17)\cite{wu2017leveraging} &6.08 (4.84) &11.54 (8.29) &6.78 (6.13) &5.73 (5.69) &5.98 (6.01) &7.33 (6.32) &6.88 (5.12) \\
			LAB(CVPR18)\cite{Wu2018LookAB} &5.27 (4.19) &10.24 (7.50) &5.51 (4.01) &5.23 (4.74) &5.15 (5.31) &6.79 (6.04) &6.32 (4.75) \\
			Wing(CVPR18)\cite{Feng2017WingLF} &5.11 (3.98) &8.75 (6.31) &5.36 (3.76) &4.93 (4.47) &5.41 (5.69) &6.37 (5.42) &5.81 (4.34) \\			
			\hline
			HGs &6.67 (5.86) &14.51 (9.44) &7.23 (6.77) &6.17 (6.04) &6.09 (6.21) &7.76 (6.54) &7.71 (5.49) \\
			HGs+MCG &5.21 (4.09) &10.09 (7.24) &5.27 (3.46) &4.93 (4.31) &4.78 (4.80) &6.39 (5.31) &6.03 (4.51) \\
			HGs+HSR &5.47 (4.33) &10.34 (7.88) &5.45 (3.87) &5.28 (4.81) &5.09 (5.17) &6.71 (6.12) &6.28 (4.81) \\
			\textbf{HGs+MCG+HSR(MHHN)} &\textbf{4.77} (3.11) &\textbf{9.31} (6.91) &\textbf{4.79} (3.13) &\textbf{4.72} (3.97) &\textbf{4.59} (4.56) &\textbf{6.17} (4.77) &\textbf{5.82} (4.28)\\
			\hline
		\end{tabular}
	\end{center}
	\label{tabwflw}
\end{table*}
\begin{table}
	\caption{The effect of different loss functions on the 300W challenging subset.}
	\begin{center}
		\begin{tabular}{p{0.7cm}p{0.7cm}p{0.7cm}p{0.7cm}p{0.7cm}p{0.7cm}p{0.7cm}}
			\hline
			Loss & MSE & MSE  +SDT &KLDL +SDT &JSDL +SDT &  SDL1 +SDT & SDL +SDT \\
			\hline
			NME(\%) &	7.56&	7.31 & 7.02 & 6.83 &6.77 &6.72 \\
			\hline
		\end{tabular}
	\end{center}
	\label{tabloss}
\end{table}
\begin{table}
	\caption{The effect of different $\sigma$ values of the HGs+HSR on the 300W challenging subset.  (- not converged)}
	\begin{center}
		\begin{tabular}{p{1.0cm}p{1.0cm}p{1.0cm}p{1.0cm}p{1.0cm}p{1.0cm}}
			\hline
			$\sigma$ & 2 & 3 &4 &5  \\
			\hline
			NME(\%) &- & 6.72 & 6.90 &7.17 \\
			\hline
		\end{tabular}
	\end{center}
	\label{tabsigmma}
\end{table}
\begin{table}
	\caption{The effect of different $\lambda$ values of the MHHN on 300W challenging subset.}
	\begin{center}
		\begin{tabular}{p{1.0cm}p{1.0cm}p{1.0cm}p{1.0cm}p{1.0cm}p{1.0cm}p{1.0cm}}
			\hline
			$\lambda$ & 1/2 & 1/4 & 1/8 &1/16 &1/32  \\
			\hline
			NME(\%) &7.01 & 6.47 & 6.21 &6.01 &6.13 \\
			\hline
		\end{tabular}
	\end{center}
	\label{tablambda}
\end{table}
\\\indent\textbf{Analyses of $\sigma$ and $\lambda$.} $\sigma$ denotes the standard deviations in the $x$- and $y$-directions of the distribution of the ground-truth landmark heatmaps, and it also determines the size of the foreground region of the landmark heatmap. In general, the value of $\sigma$ is affected by the height and width of the landmark heatmap. As our landmark heatmap size is ${\rm{64}} \times {\rm{64}}$, we conduct the corresponding experiment by using different $\sigma$ values on the 300W challenging subset. From the experimental results in Table \ref{tabsigmma}, we find that $\sigma=3$ is a good choice. $\lambda$ represents the weight coefficient of fine detection loss in the SDL. If $\lambda$ is set to 0, then it means the JSDL is used to optimize the proposed MHHN. If $\lambda$ is set to a large integer, then the MHHN is difficult to converge. Table \ref{tablambda} shows the experimental results of different $\lambda$ values on the 300W challenging subset, which indicates that the MHHN can achieve good results when  $\lambda$ is set to ${1/16}$. 
\\\indent\textbf{Search patch.} From the generated heatmaps, we can first predict the suboptimal landmarks. Then, by utilizing the SDT, we can obtain the optimal landmarks. In general, the search patch is a $9 \times 9$ rectangular region centered on the suboptimal landmark. The distribution (quality) of the heatmap can be affected by heavy occlusions and large poses. By introducing the SDL, our method can generate more effective heatmaps under complicated scenarios. Moreover, we can further expand the search radius to obtain more accurate landmarks. When we increase the search radius to $15 \times 15$, the NME of the MHHN on 300W challenging subset can be reduced from 6.01\% to 5.98\%, which further indicates that the proposed SDT can help achieve high-precision face alignment.
\\\indent\textbf{Computational cost.} Since our method can achieve heatmap subpixel accuracy, reducing the sizes of the input and output should not significantly affect the performance of our method. To verify this, we reduce the input of the MCG model from $64 \times 64 \times 256$ to  $32 \times 32 \times 256$  and the output landmark heatmap size from $64 \times 64$ to $32 \times 32$. The NME on the 300W challenging dataset only increases from 6.01\% to 6.07\%, which still outperforms state-of-the-art methods. This finding indicates that the proposed HSR method can be used to reduce computational costs in heatmap-regression-like methods without significantly degrading their performances. The MCG model also increases the computational costs. To further evaluate its effectiveness, we conduct the following experiment. We replace the last hourglass network unit of the four stacked hourglass networks with the MCG model. The NME on the 300W challenging dataset will only increase from 6.01\% to 6.13\%, which still achieves state-of-the-art accuracy. These two experiments further demonstrate the effectiveness of the MCG model and the HSR method.
\\\indent\textbf{Heatmap quality.}  Comparisons of the generated heatmaps of the GT (ground-truth), HGs, HGs+MCG+JSDL and our method are shown in Fig. \ref{fig6}. Fig. \ref{fig6} (a) illustrates some landmark heatmaps under severe occlusions. The first row, second row, third row and fourth row represent the landmark heatmaps of the ground-truth, HGs, HGs+MCG+JSDL and our method, respectively. In each row, the first four columns represent the fusion of landmark heatmaps and face images, and the last four columns represent the corresponding landmark heatmaps. From Fig. \ref{fig6} (a), we can see that HGs' heatmaps usually have long tails (the 5th, 6th and 7th heatmaps in the second row) and double centroids (the 8th heatmap in the second row) under severe occlusions, thus resulting in poor performance. Fig. \ref{fig6} (b) illustrates some landmark heatmaps (left outer mouth corner, right outer mouth corner, upper outer lip center and lower outer lip center) under large poses. From Fig. \ref{fig6} (b), we can see that all the methods can generate high-quality landmark heatmaps, but our method can obtain more accurate landmarks. As shown in Fig. \ref{fig6}, with the proposed MHHN, our heatmaps are more robust to heavy occlusions and extremely large poses, thus outperforming state-of-the-art methods.
\subsection{Ablation Study}
Our proposed multi-order high-precision hourglass network contains two pivotal components. From Table \ref{tab300w}, we can find that each proposed module plays an essential role in improving performance. The performance of the baseline HGs in Table \ref{tab300w} can achieve 4.43\% NME on the 300W Common subset and 7.56\% NME on the 300W Challenging subset. When HGs are combined with the HSR method or the MCG model, the performances of the corresponding algorithms are greatly improved (6.72\% NME and 6.51\% NME on 300W challenging subset, respectively). When integrating the HSR method and MCG model via a multi-order high-precision hourglass network, the final results of our method can be further improved (3.18\% NME on 300W Common subset and 6.01\% NME on 300W Challenging subset). The experimental results from Tables \ref{tab300w}--\ref{tabwflw} indicate that 1) the MCG model can effectively utilize multi-order cross information to enhance geometric constraints for faces under extremely large poses and heavy occlusions; 2) the SDL and SDT can be combined by the HSR method to achieve heatmap subpixel regression; and 3) the combination of the MCG model and HSR method works better than any one of them on their own, showing the complementarity of these two components.
\subsection{Experimental results and discussions}
From the experimental results listed in Tables \ref{tab300w} -- \ref{tablambda} and the figures presented in previous subsections, we state the following observations and corresponding analyses.
\\\indent (1) MHHN, HGs \cite{Yang2017StackedHN}, SAN \cite{Dong2018StyleAN} and Liu et al. \cite{Liu2019SemanticAF} are heatmap regression face alignment methods. However, from Tables \ref{tab300w} -- \ref{tabloss}, we know that the MHHN performs better than the other methods, mainly because 1) the MCG model can effectively enhance facial geometric constraints; 2) the HSR method can further help achieve high-precision facial landmark detection; and 3) by integrating the MCG model and HSR method into a multi-order high-precision hourglass network, more effective landmark heatmaps can be generated and heatmap subpixel landmark coordinates can be detected.
\\\indent (2) Both the MHHN and ODN \cite{Zhu2019RobustFL} enhance the robustness of the model to partial occlusion by capturing facial geometric constraints among different components with the matrix outer product operation. However, from the experimental results on occluded datasets, such as the 300W challenging set and the COFW dataset (see Tables \ref{tab300w} and \ref{tabcofw}), we know that our MHHN greatly exceeds the ODN, which indicates that by introducing multi-order cross information (i.e., high-order information and cross-layer information), our MCG model can obtain more discriminative representations for enhancing the geometric constraints, which leads to more robust face alignment. 
\\\indent (3) The proposed HSR method can help generate more accurate and effective landmark heatmaps by optimizing the SDL, and then, the SDT can be further used to estimate the heatmap subpixel landmark coordinates. As shown in Table \ref{tabloss}, if we replace the SDL with other loss functions or do not use the SDT, the errors on the 300W challenging set will rise significantly, which indicates that 1) fine detection loss can help generate more effective landmark heatmaps; 2) the SDT can be used to achieve heatmap subpixel landmark detection; and 3) by incorporating the SDL with the SDT, we can achieve high-precision face alignment.
\\\indent (4) As shown in Table \ref{tabloss}, when combining the MSE with the SDT, the error of face alignment on 300W challenging subset can be effectively reduced, which also indicates that the proposed SDT can not only achieve heatmap subpixel-level facial landmark detection but can also make the heatmap pixel-level corrections to the detected landmarks.
\section{Conclusion}
Unconstrained face alignment is still a very challenging topic due to the presence of large poses and partial occlusions. In this work, we present a multi-order high-precision hourglass network to address face alignment under extremely large poses and heavy occlusions. By fusing the multi-order cross geometry-aware model and the heatmap subpixel regression method with a seamless formulation, our MHHN is able to achieve more robust FLD. It is shown that the MCG model can effectively enhance geometric constraints and context information for face alignment by introducing multi-order cross information. The heatmap subpixel regression method can further improve the accuracy of face alignment by utilizing the subpixel detection loss and the subpixel detection technology, which help achieve heatmap subpixel face alignment. Experiments on four benchmark datasets of face alignment demonstrate that our method outperforms state-of-the-art methods. It can also be found from the experimental results that generating heatmaps with the distribution constraints is more effective for enhancing the accuracy and robustness of heatmap regression face alignment than generating heatmaps with pure pixel difference constraints.
\\\textbf{Acknowledgements} We thank the reviewers for the suggestions and the NSCC for computing resources.

%


\ifCLASSOPTIONcaptionsoff
  \newpage
\fi



%



\bibliographystyle{IEEEtran}
\bibliography{IEEEabrv,IEEEexample}




\end{document}